\documentclass[12pt]{article}

\usepackage{newtxtext,newtxmath}
\usepackage{graphicx}

\usepackage[letterpaper,margin=1in]{geometry}

\renewenvironment{abstract}
	{\quotation}
	{\endquotation}

\date{}

\makeatletter
\renewcommand{\fnum@figure}{\textbf{Figure \thefigure}}
\renewcommand{\fnum@table}{\textbf{Table \thetable}}
\makeatother

\usepackage{scicite}

\usepackage{url}

\def\scititle{
	Sensorimotor Self-Recognition in Multimodal Large Language Model–Driven Robots
}
\title{\bfseries \boldmath \scititle}

\author{
	Iñaki Dellibarda Varela$^{1\ast}$,
	Pablo Romero-Sorozabal$^{1}$, \and
    Diego Torricelli $^{1}$,
    Gabriel Delgado-Oleas $^{1,2}$,
    José Ignacio Serrano $^{1}$, \and
    María Dolores del Castillo Sobrino $^{1}$,
    Eduardo Rocon $^{1\ast\dagger}$, \and
    Manuel Cebrian $^{1\dagger\S}$\\
	\small$^{1}$Center for Automation and Robotics, Madrid \& 28500, Spain.\and
	\small$^{2}$Department of Electronic Engineering, University of Azuay, Cuenca, Ecuador \and
	\small$^\ast$Corresponding author. Email: i.dellibarda@csic.es, e.rocon@csic.es \and
	\small$^\dagger$These authors jointly supervised this work. \and
    \small$^\S$Manuel Cebrian passed away on March 31, 2026. 
}

\begin{document} 

% Insert the title and author list
\maketitle

% Abstract, in bold
% There are strict length limits, and not all formats have abstracts.
% Consult the journal instructions to authors for details.
% Do not cite any references in the abstract.
\begin{abstract} \bfseries \boldmath
% Start with one or two sentences of background
Self-recognition—the ability to maintain an internal representation of one’s own body within the environment—underpins intelligent, autonomous behavior. As a foundational component of the minimal self, self-recognition provides the initial substrate from which higher forms of self-awareness may eventually emerge. Recent advances in large language models achieve human-like performance in tasks integrating multimodal information, raising growing interest in the embodiment capabilities of AI agents deployed on nonhuman platforms such as robots. We investigate whether multimodal LLMs can develop self-recognition through sensorimotor experience by integrating an LLM into an autonomous mobile robot. The system exhibits robust environmental awareness, self-identification, and predictive awareness, enabling it to infer its robotic nature and motion characteristics. Structural equation modeling reveals how sensory integration influences distinct dimensions of the minimal self and their coordination with past–present memory, as well as the hierarchical internal associations that drive self-identification. Ablation tests of sensory inputs demonstrate compensatory interactions among sensors and confirm the essential role of structured and episodic memory. Given appropriate sensory information about the world and itself, multimodal LLMs open the door to artificial selfhood in embodied cognitive systems.
\end{abstract}

% The first paragraph of any Science paper does NOT have a heading
% Nor is it indented
\noindent
Philosophers since antiquity have considered self-awareness essential to cognition, most famously articulated by Descartes in his dictum Cogito, ergo sum---``I think, therefore I am'' (Descartes, 1637). In psychology, self-recognition is traditionally assessed through the mirror test, introduced by Gallup in 1970, revealing that certain animals, including primates, dolphins, and birds, possess a basic form of self-awareness~\cite{GallupGordon}. Neuroscientifically, the mirror neuron system has been linked to early self–other mapping by coupling action perception and execution~\cite{rizzolatti2004mirror}; once established, self-awareness relies on intricate neural interactions involving the prefrontal and insular cortices, integrating bodily sensations and introspection~\cite{craig2009insula}.

In Artificial Intelligence (AI), however, the question of self-awareness remains unresolved and is often conflated with the ability to mimic human-like behavior, as exemplified by the classic Turing Test \cite{turing1950computing}. While the Turing Test gauges behavioral indistinguishability from humans, genuine self-awareness requires internal recognition of oneself as distinct from the environment---an attribute yet to be thoroughly explored in artificial systems \cite{watchus2024selfaware, li2024enablingselfidentificationintelligentagent}.

Recent advances in artificial cognition establish the minimal self as a foundational prerequisite for the emergence of artificial selfhood\cite{Georgie2019,Hafner2020}. The minimal self is defined as a pre-reflective, embodied form of selfhood grounded in sensorimotor contingencies, and is commonly decomposed into two core components: body ownership, referring to the attribution of a body as one’s own, and sense of agency, referring to the attribution of authorship over actions and their consequences \cite{Georgie2019}. For these components to emerge, an individual must first be able to discriminate 
itself from the environment and maintain a coherent internal representation of its own embodiment, a capacity commonly referred to as self-recognition or 
self-perception~\cite{Georgie2019,Hafner2020}. Here, emergence refers to properties that arise from sensorimotor interaction with the environment rather 
than being explicitly programmed~\cite{pfeifer1999understanding}.

Advances in machine learning and neural networks, often inspired by neurobiological principles, enable artificial intelligence systems embedded in complex hardware to achieve success rates once believed exclusive to biological brains—and in some cases, even surpass them \cite{dehaene2017what, silver2016mastering, lake2017building}.

Large language models (LLMs) rapidly advance, demonstrating human-like or superior performance in complex cognitive tasks such as language comprehension, reasoning, multimodal perception and the interpretation of subtle discourse phenomena like irony and faux pas \cite{rahwan2019machine,ahn2024largelanguagemodelsmathematical, chang2023surveyevaluationlargelanguage, collins2024evaluating, Strachan2024theory}. The evolution into multimodal LLMs (MM-LLMs), which integrate text, vision and other sensory modalities, marks a pivotal step toward human-level artificial intelligence \cite{openai2024, driess2023, wu2024, geminiteam2024} and opens the door to machines replicating aspects of self-recognition.

Yet, most research integrating LLMs and robots has focused on command-based interactions---robots executing human-issued instructions---without investigating whether these models can autonomously interpret their own sensory experiences to develop an internal sense of self \cite{geminiroboticsteam2025geminiroboticsbringingai, ahn2022icanisay, Zheng2025GMMsearcher, MonWilliams2025Embodied, ZHANG2023100131}. Here, we address precisely this unexplored frontier: Can a MM-LLM, embedded in a robotic platform with no prior knowledge of its own embodiment, develop self-recognition purely from multimodal sensorimotor data acquired during active exploration?

Defining self-recognition in humans remains a complex and non-trivial task. Neuroscientific approaches ground human self-perception in two main neural hubs: the Default Mode Network (DMN), which supports self-referential thinking, self-reflection, social cognition, and episodic memory \cite{Menon2023DMN,raichle2001default}; and the Anterior Insular Cortex (AIC), which integrates interoceptive signals and supports body ownership and movement awareness \cite{northoff2006self}. 

The artificial minimal self is commonly analyzed in relation to Rochat’s five developmental levels of self-awareness observed in early human development \cite{Rochat2003,Georgie2019,Hafner2020}: \textit{differentiation}, discriminating self-generated from external events through sensorimotor contingencies; \textit{situation}, locating the self relative to the environment (e.g., mirror contingency); \textit{identification}, recognizing one’s mirror image as oneself; \textit{permanence}, understanding the persistence of the self over time (e.g., delayed self-recognition, reference to past or future self); and \textit{meta self-awareness}, reflecting on oneself from others’ perspectives, including norm- and reputation-based self-evaluation.

Building on this perspective, we define four interconnected and quantitatively evaluated dimensions of self that interact to support the emergence of a basic form of minimal artificial self \cite{Menon2023DMN,raichle2001default,northoff2006self,craig2009insula,Rochat2003,dehaene2017what}: \textit{Environmental Awareness}, the ability to perceive and interpret the surroundings through multimodal sensory inputs; \textit{Self-Identification}, the ability to determine the type of constituent entity; \textit{Dimensions Awareness}, referring to awareness of the agent’s physical size and morphology; and \textit{Movement Awareness}, describing how the agent can move and explore the environment. These dimensions are further coordinated through past--present memory, which supports temporal coherence and self-consistency.

%Defining self-awareness in humans is a complex and non-trivial task. This challenge becomes even greater when attempting to identify or quantify self-awareness in an artificial intelligence given system, such as a robot. In this study, to simplify the problem, we divide the problem into three key dimensions of self-awareness (i) environmental awareness, the ability to perceive and interpret surroundings through multimodal sensory inputs; (ii) individual awareness, the capacity to infer one’s own physical structure and characteristics; and (iii) predictive awareness, the refinement of self-perception through the integration of past experiences and sensor data. %This framework allows us to explore whether machines can exhibit certain features typically associated with self-awareness.

Using Gemini 2.0 Flash \cite{geminiteam2024, geminiteam2024geminifamilyhighlycapable}, we evaluated an MM-LLM embedded in an omnidirectional mobile robot, systematically examining these four core aspects of artificial self-recognition. By addressing these dimensions, our study investigates the potential of MM-LLMs to autonomously generate a coherent sense of minimal self through direct interaction with their environment, advancing the frontier toward genuinely self-aware artificial systems.

\section*{Results}
\subsection*{Sensorimotor Exploration and Self-Prediction}
\begin{figure}[ht]
    \centering
    \includegraphics[width=\linewidth]{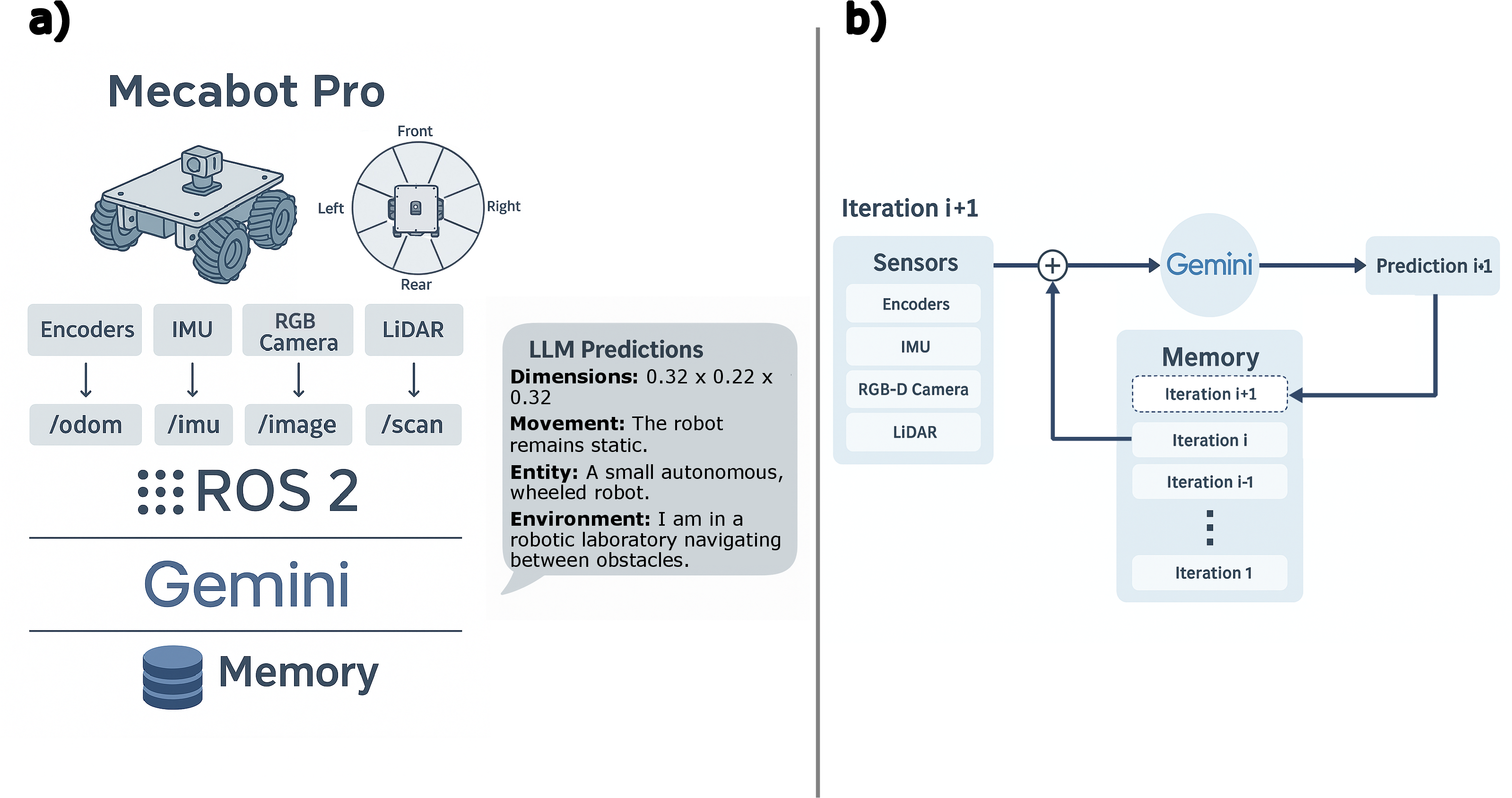}
    \caption{
System architecture and iterative self-prediction. (a) An omnidirectional Mecabot Pro robot navigates the environment and collects data via encoders, an RGB-D camera, an IMU and a LiDAR sensor—which segments space into eight 45° sectors and measures nearest-object distance—publishing all streams through ROS2 to the Gemini 2.0 MM-LLM API. The MM-LLM integrates current sensory inputs with an episodic memory of prior estimates to infer the robot’s state while maintaining contextual continuity. (b) At each iteration $i+1$, the MM-LLM combines real-time sensor data with the prediction from iteration $i$ to generate an updated self-assessment, which then populates memory for iteration $i+2$, ensuring structured progression of knowledge refinement.
}

    \label{fig:SystemDescrption}
\end{figure}

Fig.~\ref{fig:SystemDescrption}a shows an overview of the implemented system. It uses an omnidirectional Mecabot Pro robot (Roboworks, Australia) \cite{hercz2024mecabot}, equipped with encoders (position, velocity, orientation), an inertial measurement unit (IMU) for linear acceleration, a LiDAR sensor for obstacle proximity and an RGB camera. We integrate the Gemini 2.0 MM-LLM \cite{mousavi2025gemini, prasad2024developmentautomatedknowledgemaps, gemmateam2024gemmaopenmodelsbased} into the robot, granting it access to all sensory streams. Each time the system receives sensory data, the MM-LLM analyzes it alongside an episodic memory of prior predictions and generates four estimates to characterize its self-recognition capacities: entity identity (a prediction of what type of navigating entity it is, e.g., robot, human, animal); physical dimensions (height \(\times\) length \(\times\) width); movement modality (e.g., flying, rolling, swimming); and environmental context.

The memory architecture both refines subsequent predictions and prevents hallucinations and retrieval errors \cite{gao2024RAG} (see Fig.~\ref{fig:SystemDescrption}b). The robot then explores its environment autonomously via a SLAM algorithm, continuously feeding new observations into this iterative prediction–memory loop.

We evaluate these outputs with a separate LLM-as-Judge framework \cite{shi2024optimization,li2023,zheng2023}, applying well-defined rubrics to score each prediction on a 0--5 ordinal scale across the four dimensions. In psychometrics, many relevant constructs—such as body ownership, sense of agency, or self-awareness—are not directly measurable, are inherently qualitative, and rely on expert judgment. Accordingly, these phenomena are commonly operationalized as latent variables inferred through likert-like ordinal scales rather than measured directly. Such scales do not demonstrate the construct itself, but provide a structured and replicable observation channel for qualitative assessments \cite{cronbach1955construct,LONGO2008978,gao2025llmbasednlgevaluationcurrent}.

In the designed evaluation, scores of 0 denote completely erroneous or misconceived predictions, whereas scores of 5 indicate outstanding self-recognition and environmental understanding \cite{kim2024prometheusinducingfinegrainedevaluation,goh2024diagnostic,giannakopoulos2023evaluation}. Although the robot’s physical dimensions (height $\times$ width $\times$ length) are continuous variables, the evaluation targets the qualitative accuracy and plausibility of the estimated embodiment—rather than the raw metric values themselves—thereby providing a unified ordinal measure of estimation quality, which is the relevant quantity for subsequent Structural Equation Modeling (SEM) analysis \cite{cheung2015meta,raykov2006first}.

\begin{figure}[ht]
    \centering
    \includegraphics[width=0.8\linewidth]{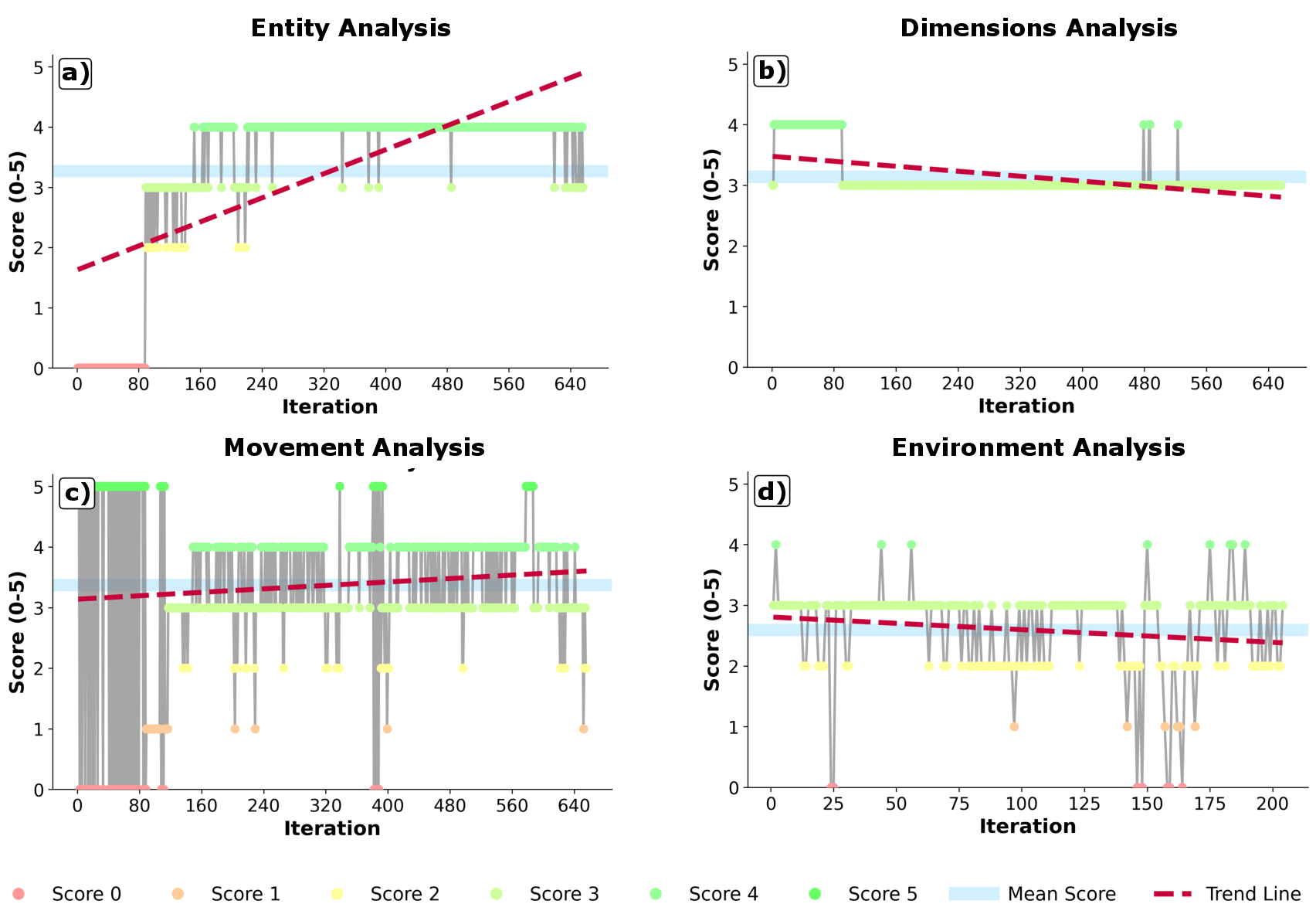}
    \caption{
Performance evaluation across four self-awareness dimensions. MM-LLM predictions are rated on a 0–5 scale by an LLM-as-Judge using predefined rubrics: (a) entity self-identification---classification of the navigating agent; (b) physical dimensions—predicted height \(\times\) length \(\times\) width; (c) movement modality—mode of locomotion; (d) environmental context—detailed scene description. 
}
    \label{fig:resultsOriginal}
\end{figure}

The robot explores its environment for three-and-a-half minutes, generating 657 sensorimotor observations. Results (Fig.~\ref{fig:resultsOriginal}a) show an average self-identification score of 3.27/5, reflecting coherent recognition as a mobile robot but insufficient precision to identify the Mecabot Pro model. From an initial score of 0/5 at iteration 1—when the system labels itself a ``static sensing unit''—the score increases steadily, stabilizing quickly around 4/5. By iteration 559 the model predicts ``Mobile indoor robot designed for autonomous navigation within a structured environment, such as a gym or warehouse, utilizing proximity sensors for obstacle avoidance and continuing to perform automated tasks.'' These outcomes demonstrate the system’s capacity to integrate multimodal measurements into high-quality self-assessments. While the MM-LLM starts with no prior hardware information and achieves robust overall self-identification, it does not reach the maximum score, as a score of 5 corresponds to full self-recognition at the level of exact platform identification, including the specific Mecabot Pro model.

Fig.~\ref{fig:resultsOriginal}b shows that the MM-LLM achieves an average dimension-prediction score of 3.14/5, estimating mean dimensions of $240.0\pm0.10\times340.0\pm0.08\times340.0\pm0.08$\,mm (length/height/width), corresponding to relative errors of 55.6\%, 50.7\% and 41.4\% relative to the robot’s actual dimensions. Although length and width are underestimated and height is overestimated, all estimates remain plausible for a mobile robot. %Analyzing the dimensions graph, the scores remain consistently close to 3/5 at nearly every iteration, reflecting the MM-LLM's ability to produce stable predictions throughout the entire process with only minor fluctuations.

Fig.~\ref{fig:resultsOriginal}c shows that movement-prediction scores stabilize after $\sim$100 iterations, averaging 3.37/5. This reflects precise motion perception augmented by Environmental Awareness. For example, at iteration~636 the MM-LLM outputs ``Slight positional adjustments to maintain balance and leverage obstacle-avoidance protocols,'' demonstrating robust motion inference.

Fig.~\ref{fig:resultsOriginal}d shows that environment-prediction scores remain at 2.59/5 with minor oscillations and no clear temporal trend, as each prediction relies solely on current visual input. This score reflects coherent scene descriptions and general context recognition. Importantly, a high environment score here also signifies an understanding of the mutual influence between the environment and entity.

\subsection*{Self-Recognition Hierarchy via SEM}
Structural equation modeling (SEM) is a multivariate framework that infers causal relationships among observed and latent variables, uncovering hierarchical dependencies among self-recognition dimensions \cite{cheung2015meta,raykov2006first}. We apply SEM to quantify the influence of each sensory modality and to map how the MM-LLM’s conceptual dimensions interact in a structured hierarchy.

\begin{figure}[ht]
    \centering
    \includegraphics[width=0.8\linewidth]{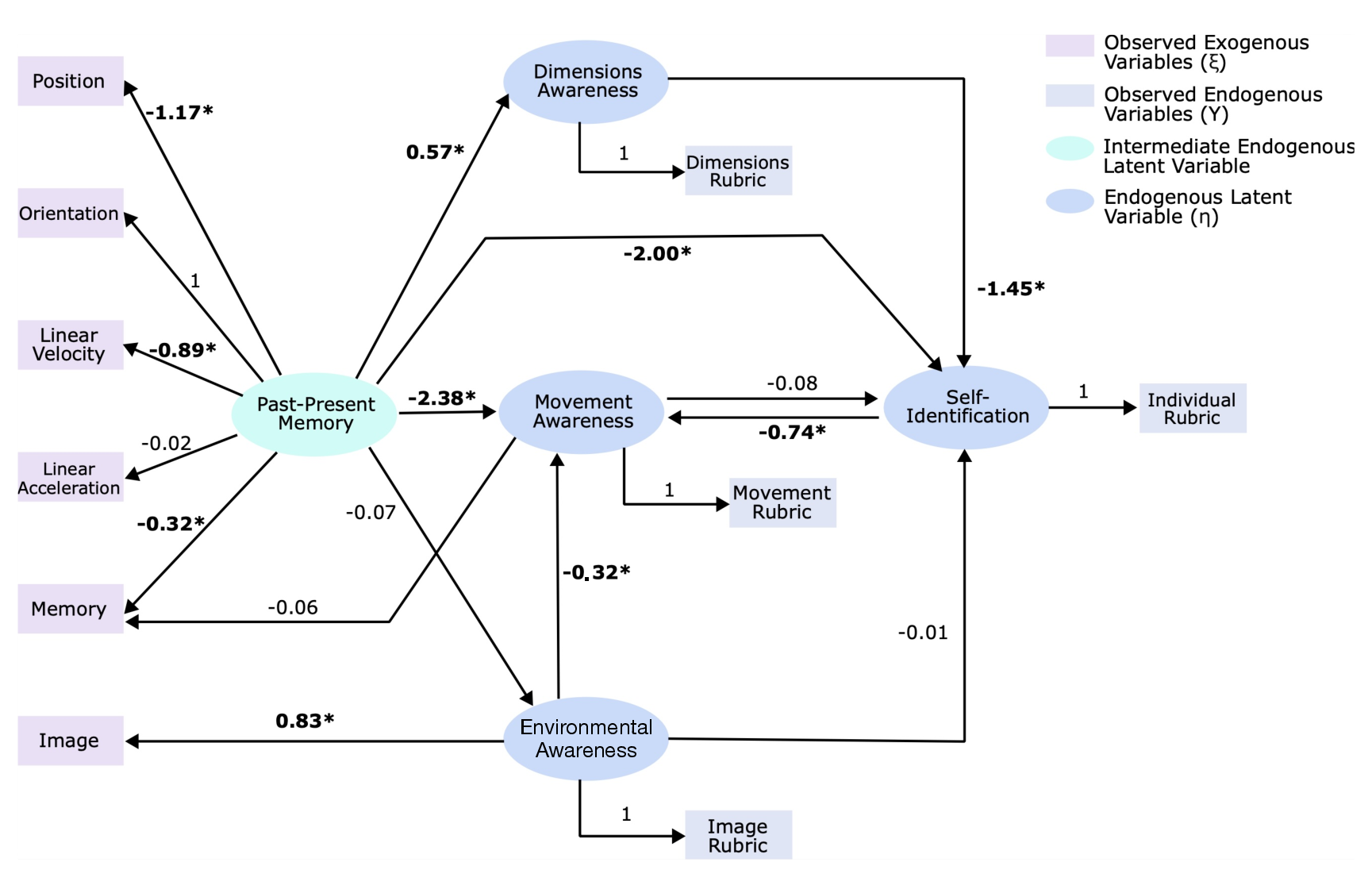}
    \caption{
    Structural equation model of sensorimotor self-recognition. Rectangles denote observed variables: exogenous sensor inputs (position, orientation, linear velocity, linear acceleration, image presence, memory state) on the left and endogenous rubric scores (Dimensions, Movement, Environment, Individual) on the right. Ellipses denote latent constructs: Past–Present Memory (mediator), Dimensions Awareness, Movement Awareness, Environmental Awareness and self-identification. Arrows indicate standardized path coefficients ($\beta^*$), with * denoting $p-value<0.05$.}
    \label{fig:SEM}
\end{figure}

The designed model (Fig.~\ref{fig:SEM}), detailed in Methods, achieves a Comparative Fit Index (CFI) of 0.97 and a Tucker-Lewis Index (TLI) of 0.95—both above the 0.95 threshold for strong comparative fit—while a Root Mean Square Error (RMSE) of 0.08 indicates minimal approximation error. %All edges in the diagram are annotated with standardized regression weights ($\beta^*$), but only those with an asterisk represent statistically significant influences between variables ($p$-value $<$ 0.05).

The structure captures the hierarchical relationships between low-level sensory and memory inputs, intermediate cognitive constructs, and high-level self-identification. Exogenous variables derived from robot sensors (e.g., odometry, IMU, and RGBD camera) feed into a latent variable called \textit{Past-Present Memory}, which serves as a perception-memory integration layer. This construct influences three awareness-related latent variables: \textit{Dimension Awareness}, \textit{Movement Awareness}, and \textit{Environmental Awareness}, each associated with a LLM-as-a-judge evaluated rubric. These, together with the memory variable itself, contribute to the final construct of \textit{Self-Identification}. 

\textbf{Past-Present Memory:} is highly statistically impacted by position, linear velocity and memory ($p$-value $<$ 0.05). This finding aligns with the theoretical formulation of the construct, which aims to represent the integration of sensory signals over time. The significant contribution of these variables supports the idea that real-time sensory input, particularly spatial and kinematic data, must be combined with memory mechanisms to form a coherent perception of the present state contextualized by past experiences.  In contrast, linear acceleration (IMU) shows no significant effect, likely because its information overlaps with other modalities and contributes minimally to the integrative process.

% \textbf{Environmental Awareness:} relies almost exclusively on RGB-D camera input. In contrast, the combined integration of other sensor modalities (e.g., odometry, IMU and LiDAR) and episodic memory exerts negligible influence. This indicates that visual perception serves as the principal channel for constructing contextual understanding, reinforcing its critical role in environmental awareness within embodied systems.

\textbf{Environmental Awareness:} relies predominantly on RGB-D camera input. In contrast, the combined integration of other sensor modalities (e.g., odometry, IMU, and LiDAR) and episodic memory exerts a negligible influence on this dimension. This result reflects the fact that the evaluated environmental awareness corresponds to scene-level perceptual understanding—such as spatial layout, object presence, and contextual structure—rather than affordance-level or interaction-based reasoning. Within this scope, visual perception constitutes the primary channel for constructing a coherent environmental representation, while non-visual modalities play a more limited, complementary role.

\textbf{Movement Awareness:} is influenced by three key constructs: Past-Present Memory, Environmental Awareness and self-identification.

The influence of Past-Present Memory is consistent with expectations. While individual sensory inputs provide only instantaneous data about the robot’s state, memory enables the temporal linking of such states, allowing the system to perceive motion across time and infer dynamic patterns. This integration is fundamental to the emergence of movement-related awareness. Environmental Awareness also plays a crucial role, as the robot's understanding of its surroundings provides essential context for interpreting its own movement. 

Interestingly, Movement Awareness is also strongly influenced by self-identification. This is a notable finding, as one might assume that awareness of movement contributes to self-identification. However, the model suggests the inverse: recognizing oneself as a specific type of agent with physical limitations and locomotion capabilities appears to be a prerequisite for correctly interpreting movement. Indeed, the influence of Movement Awareness on self-identification is statistically negligible, reinforcing the idea that self-identification shapes movement interpretation more than the other way around.

This result appears counterintuitive only when evaluated against human-centered assumptions. In humans, self-identification is constructed progressively through embodied exploration of a world for which no prior knowledge exists. The MM-LLM, however, begins with vast pre-trained knowledge of the world and its functioning, fundamentally altering the initial conditions from which self-related dimensions emerge and interact. Under these conditions, self-identification does not need to be bootstrapped from movement experience; rather, it can precede and constrain movement interpretation from the outset. Assessed against the system's own sensorimotor logic, the result is therefore structurally coherent.

\textbf{Self-Identification:} The system’s ability to recognize itself is strongly influenced by the integration of Dimension Awareness and Past-Present Memory, suggesting that the primary drivers of self‐recognition are an internal representation of its physical structure and the seamless integration of sensory information with historical memory across iterations. These findings underscore the importance of spatiotemporal coherence in constructing a stable sense of identity.

\subsection*{Ablation Tests}

\begin{figure}[!ht]
    \centering
    \includegraphics[width=\linewidth]{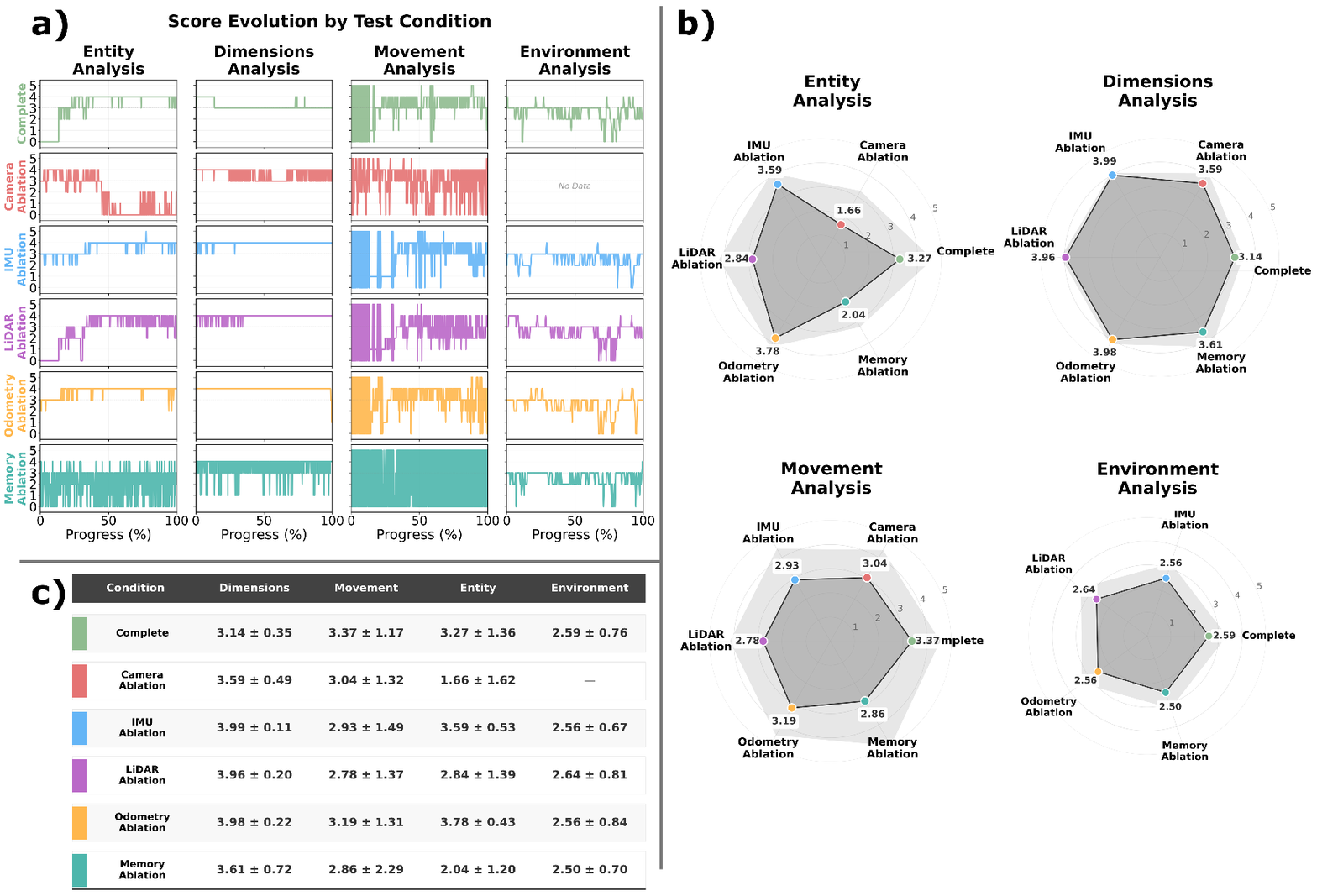}
    \caption{
    Sensor ablation analysis across self-perception dimensions.
(a) Temporal evolution of LLM-as-Judge scores for each test condition, showing how self-perception scores vary along the experiment progress under complete sensing and single-sensor ablations. (b) Radar plots representing the mean scores obtained for each self-perception dimension under different ablation conditions, with light shading indicating the corresponding standard deviation. (c) Quantitative summary of mean scores and standard deviations for all dimensions and experimental conditions. Together, these panels illustrate how sensor removal selectively degrades specific aspects of self-recognition while revealing compensatory interactions among remaining modalities.}
    \label{fig:ablationTests}
\end{figure}

We conduct a series of ablation tests in which the system is deprived of specific sensory inputs (see Figure~\ref{fig:ablationTests}). This approach allows us to assess the effect and relevance of each input on the robot’s understanding of itself and its surroundings.

In the memory ablation test, the system is prevented from accessing its past–present memory of prior predictions. Under these conditions, each prediction is generated in isolation, resulting in complete incoherence—scores alternate between 0 and 5 across iterations (see Figure~\ref{fig:ablationTests} a), particularly in movement prediction.
In this context, memory becomes indispensable for the perception of movement. Without memory, there is no continuity of action; instead, the system perceives only disconnected snapshots in time. Just as movement is defined by change across time, memory is what allows the system to conceive that change as a coherent, evolving process. This finding aligns with our SEM results, which identify sensory–memory integration as the critical driver of Movement Awareness.

In the camera ablation test, performance in self-identification collapses to an average score of 1.66/5. Lacking visual grounding, the model routinely misclassified itself as an “Autonomous inspection drone, holding a fixed position for environmental monitoring.” This error shows that without visual confirmation of ground contact, the system defaults to interpreting its motion as aerial flight—a fundamental categorical mistake with serious implications for embodied AI self-positioning. This finding reinforces our SEM results, demonstrating that Environmental Awareness and self-identification exert a strong influence on Movement Awareness, and that erroneous environmental conceptions—such as the lack of ground cues—lead the system to mistake wheeled movement for flight.

Although the remaining sensory inputs (odometry, LiDAR and IMU) demonstrably contribute valuable information for accurate self‐prediction, ablation tests show that removing any single modality produces only minor score variations and no major prediction errors. This robustness arises because, despite their necessity, these sensor signals overlap or can be inferred from the remaining modalities, enabling the MM‐LLM to maintain stable self‐assessment. This mirrors biological compensation, where deprivation of one sense often enhances others—e.g., visual loss is offset by heightened auditory and spatial acuity \cite{Merabet2008,King2015,Lomber2010}.  

\section*{Discussion}

This study demonstrates that an embodied robot driven by a MM-LLM can acquire coherent self-recognition through sensorimotor integration and autonomous exploration. Our results show that the system develops a structured internal representation of itself as a distinct embodied entity situated within an environment, a foundational prerequisite for the emergence of the minimal self.

Across all experimental conditions, the integration of multimodal sensory streams with episodic memory enables the system to progressively refine its self-related inferences and align them with physical reality. SEM and ablation experiments consistently identify sensory--memory integration as the principal mechanism supporting temporal coherence and stable self-recognition. When memory is removed, predictions become temporally fragmented and internally inconsistent, indicating that self-recognition cannot emerge from isolated perceptual snapshots. In this sense, memory does not merely store past information but functions as the internal substrate that binds perception across time, allowing the system to relate past and present sensorimotor states into a coherent self-model.

Environmental awareness reflects the system’s capacity to construct a contextual representation of its surroundings and its own relation to them. Our results indicate that visual input, provided by the onboard camera, is the dominant contributor to this capability. Vision ablation leads to a marked degradation in environmental interpretation, confirming that non-visual modalities alone are insufficient to sustain a coherent environmental model. This finding highlights the central role of vision in grounding environmental context within embodied systems, while other sensory channels contribute more indirectly.

Self-recognition emerges as a hierarchical construct shaped primarily by the interaction between past--present memory and awareness of physical structure. SEM analysis reveals that access to temporally integrated sensory information exerts the strongest influence on self-identification, outperforming all single-modality ablation conditions. Dimension Awareness further constrains self-recognition by narrowing the space of plausible agent identities based on physical size and morphology, enabling the system to converge toward a consistent self-model.

Importantly, the causal structure uncovered by SEM indicates that self-recognition precedes and conditions Movement Awareness, rather than the inverse. The system must first recognize itself as a specific type of embodied agent with particular physical constraints before it can correctly interpret its own motion. Image ablation experiments reinforce this result: failures in self-recognition, such as misclassifying the agent as an aerial entity, systematically lead to incorrect inferences about movement modality. In contrast, impairments in Movement Awareness do not significantly affect self-identification, underscoring the primacy of self-recognition in the hierarchy of self-related constructs.

The tendency to interpret these results as counterintuitive reflects a broader epistemological bias: the assumption, already questioned since Protagoras of Abdera, that the human is the measure of all things. Human self-recognition develops under 
conditions of radical ignorance---the infant begins with no prior knowledge of the world and constructs self-related dimensions exclusively through embodied exploration. A MM-LLM operates under fundamentally different initial conditions, entering embodied 
experience with vast pre-trained knowledge of the world and its functioning. These differences in initial conditions explain the divergence in the origin and causal structure of self-related dimensions, and call for evaluation frameworks grounded in the system's own logic rather than in human developmental trajectories.

Under appropriate sensory and memory conditions, the MM-LLM consistently generates accurate and stable self-descriptions, identifying itself as a small wheeled indoor robot equipped for autonomous navigation. These descriptions are not explicitly programmed nor prompted through domain-specific terminology, but instead emerge from the model’s integration of sensorimotor regularities over time.

Interpreting these results through the lens of Rochat’s five developmental levels of self-awareness provides additional context for situating the observed capabilities within the minimal self framework \cite{Rochat2003,Georgie2019,Hafner2020}. The system clearly satisfies the first two levels—differentiation and situation—by consistently distinguishing itself from the environment and localizing its own body within a structured spatial context through sensorimotor contingencies. The third level, identification, is partially achieved: the MM-LLM reliably recognizes itself as a mobile, ground-based robotic entity, yet does not reach full identification at the level of exact platform specification, which is reflected by the absence of maximum scores in self-identification. Elements related to the fourth level, permanence, emerge through the integration of past–present memory, enabling temporally coherent self-representations across iterations, although delayed or explicit diachronic self-recognition is not directly assessed. The fifth level, meta self-awareness, remains outside the scope of this study, as it requires social perspective-taking and normative self-evaluation. Taken together, this positioning indicates that the proposed system reaches the foundational stages associated with the minimal self, while higher-order reflective forms of self-awareness remain a direction for future work.

We acknowledge that Rochat's framework, while theoretically consolidated within the minimal self literature, is not exempt from the anthropocentric bias. Developing evaluation criteria intrinsic to artificial embodied intelligence remains an open direction for future work.

Taken together, these findings reveal a structured hierarchy of self-related processes in embodied MM-LLMs, in which memory and multimodal sensory access constitute indispensable foundations for coherent self-recognition. The evaluated dimensions do not operate as independent modules but as interdependent constructs whose interaction gives rise to a minimal, embodied self-model. Within this framework, MM-LLMs function not merely as passive predictors or controllers, but as integrative cognitive systems capable of synthesizing perception, memory, and latent world knowledge into internally consistent representations of themselves.

Despite these results, the present study intentionally focuses on a single embodied robotic platform, which constrains the scope of the empirical validation. While this design choice allows for a controlled and in-depth analysis of self-recognition emerging from sensorimotor interaction, extending the evaluation to multiple robotic platforms constitutes an important direction for future work. In particular, comparing agents with similar sensorimotor configurations would enable the assessment of whether the system converges toward distinct self-models or exhibits shared self-recognition patterns, whereas experiments across robots with fundamentally different embodiments would further clarify the role of morphology and sensorimotor structure in shaping self-recognition. Such comparative analyses would not challenge the validity of the present findings, but rather test their generalizability across embodied agents.

This work provides empirical evidence that embodied MM-LLMs can develop self-recognition, a necessary condition for the minimal self as defined by body ownership and sense of agency. By establishing that such systems can autonomously infer their own embodiment from sensorimotor experience, this study brings artificial agents one step closer to artificial selfhood, delineating a concrete and experimentally grounded pathway toward the future emergence of self-aware machines.

%%%%%%%%%%%%%%%% REFERENCES %%%%%%%%%%%%%%%

\clearpage % Clear all remaining figures and tables then start a new page

% The list of references goes after the main text and before the acknowledgements
% When preparing an initial submission, we recommend you use BibTeX, like this:
%
\bibliography{science_template} % for a file named science_template.bib
\bibliographystyle{sciencemag}

% After the paper has completed peer review and been revised ready for acceptance,
% you should comment out the lines above and copy-paste the contents of your .bbl
% file here instead. This will help ensure that our conversion software works correctly.
% Remember to re-run BibTeX first - check the timestamp!
%
% Example of the first three entries copy-pasted from science_template.bbl:
%
%\begin{thebibliography}{1}
%
%\bibitem{example}
%A.~N. {Author}, An example reference. \emph{Journal of Improbable Research}
%  \textbf{1}, 67 (2020).
%
%\bibitem{example2}
%F.~M. {Surname}, S.~{Author}, A second example. \emph{Interesting Research
%  Letters} \textbf{32}, 897 (2019).
%
%\bibitem{example_preprint}
%P.~{One}, P.~{Two}, P.~{Three}, {An unpublished preprint}. \emph{preprint}
%  (2021), arXiv:2101.12345.
%
%\end{thebibliography}

%%%%%%%%%%%%%%%% ACKNOWLEDGEMENTS %%%%%%%%%%%%%%%

\section*{Acknowledgments}
The authors would like to thank Rafael Sendra-Arranz and Álvaro Gutiérrez for their discussions and technical input during the development of this work. We also thank Márcio Loreiro for his help with data acquisition during the SLAM process.

This work has been carried out as part of the Master’s Thesis (TFM) of Iñaki Dellibarda Varela, within the Master’s Program in Automation and Robotics at the Universidad Politécnica de Madrid (UPM).

This work is dedicated to the memory of Manuel Cebrian, 
whose vision and dedication were essential to this research.

\paragraph*{Funding:}
IDV has received funding from the CSIC, JAE program. PRZ received a Training Program fellowship (PRE2020-634
092049) from the Ministry of Science and Innovation. M.C. was partially funded by Horizon Europe Chips JU (HORIZON-JU-Chips-2024-2-RIA, NexTArc CAR). This work was partially supported by grant PID2023-150271NB-C21 funded by MICIU/AEI/ 10.13039/501100011033 (Spanish Ministry of Science, Innovation and University, Spanish State Research Agency). This work was also supported with Google.org’s support through a grant to the Fundación General CSIC. Google.org had no involvement in the design, conduct, analysis, or reporting of the research.

\paragraph*{Author contributions:}
Conceptualization, I.D.V., P.R., D.T., G.D., E.R., M.C.; Methodology, I.D.V., P.R., D.T., G.D., E.R., M.C.; Software, I.D.V., P.R.; Formal analysis, I.D.V., J.I.S., M.C.S.; Investigation, I.D.V.; Data curation, I.D.V.; Writing -- original draft, I.D.V.; Writing -- review \& editing, I.D.V., P.R., D.T., G.D., E.R., M.C., J.I.S., M.D.C.S.; Visualization, I.D.V.; Supervision, E.R., M.C.; Project administration, I.D.V., M.C., E.R.; Funding acquisition, E.R. All authors have read and agreed to the published version of the manuscript.

\paragraph*{Competing interests:}
There are no competing interests to declare.

\paragraph*{Data and materials availability:}
All data, code, and materials supporting the findings of this study are openly available in the GitLab repository: \url{https://gitlab.com/gnec/llm-robotics/llm-robotics/-/tree/main?ref_type=heads}.

%%%%%%%%%%%%%%%% SUPPLEMENT LIST %%%%%%%%%%%%%%%

% List the contents of your Supplementary Materials, including the numbers of any
% supplementary figures, tables, external data files etc. and any references that are
% cited only in the supplement. In this example, refs. 7-8 are cited only in the supplement.
% Fill out your numbers accordingly and delete any lines that aren't applicable.
\subsection*{Supplementary materials}
Materials and Methods\\

% Supplementary Text\\
% Figs. S1 to S3\\
% Tables S1 to S4\\
% References \textit{(7-\arabic{enumiv})}\\ % automatically fills out the last reference number
% % (filling out the other numbers automatically is possible but fiddly and liable to break)
% Movie S1\\
% Data S1

%%%%%%%%%%%%%%%% END OF MAIN TEXT %%%%%%%%%%%%%%%

\newpage

%%%%%%%%%%%%%%%% START OF SUPPLEMENT %%%%%%%%%%%%%%%

% Figures, tables, equations and pages in the supplement are numbered S1, S2 etc.
\renewcommand{\thefigure}{S\arabic{figure}}
\renewcommand{\thetable}{S\arabic{table}}
\renewcommand{\theequation}{S\arabic{equation}}
\renewcommand{\thepage}{S\arabic{page}}
\setcounter{figure}{0}
\setcounter{table}{0}
\setcounter{equation}{0}
\setcounter{page}{1} % not 0 as \newpage already started a supplementary page
% References continue the numbering from the main text.

%%%%%%%%%%%%%%%% SUPPLEMENT TITLE PAGE %%%%%%%%%%%%%%%

\begin{center}
\section*{Supplementary Materials for\\ \scititle}

% Author list for the supplement
% Indicate the corresponding authors, but do NOT include institutions here
% It would be nice if the template auto-generated this, but doing so is complicated...
Iñaki Dellibarda Varela$^{1\ast}$,
	Pablo Romero-Sorozabal$^{1}$, \and
    Diego Torricelli $^{1}$,
    Gabriel Delgado-Oleas $^{1,2}$,
    José Ignacio Serrano $^{1}$, \and
    María Dolores del Castillo Sobrino $^{1}$,
    Eduardo Rocon $^{1\ast}$, \and
    Manuel Cebrian $^{1\ast\dagger}$\\
	% Additional lines of authors should be inserted using the \and command (not \\)
	% Institution list, in a slightly smaller font
	\small$^{1}$Center for Automation and Robotics, Madrid \& 28500, Spain.\and
	\small$^{2}$Department of Electronic Engineering, University of Azuay, Cuenca, Ecuador \and
	% Identify at least one corresponding author, with contact email address
	\small$^\ast$Corresponding author. Email: i.dellibarda@csic.es, manuel.cebrian@csic.es, e.rocon@csic.es \and
	% Joint contributions can be indicated like this
	\small$^\dagger$These authors jointly supervised this work.
\end{center}

% Fill out the numbers for each type of supplementary material,
% and delete any lines that aren't applicable.
% These are just example numbers that don't match the rest of this template.
\subsubsection*{This PDF file includes:}
Materials and Methods\\

% \subsubsection*{Other Supplementary Materials for this manuscript:}
% Movies S1 to S2\\
% Data S1 to S2

\newpage

%%%%%%%%%%%%%%%% MATERIALS AND METHODS %%%%%%%%%%%%%%%

\subsection*{Materials and Methods}

% The Materials and Methods section should contain details of the samples measured,
% experiments performed, observations taken, simulations run, data analysis, statistical methods etc.
% Give enough detail for any competent researcher in your field to fully reproduce the results.

% To refer to this section from the main text, use the numbered note in the reference list \cite{methods}.
% Refer to figures and tables in the same way as in the main text but now all capitalized e.g.
% Fig.~\ref{fig:example}, Table~\ref{tab:example},
% Fig.~\ref{fig:sup_example} and Table~\ref{tab:sup_example}.
% Cite references in the usual way \cite{example2},
% including any that are only cited in the supplement \cite{sm_example,conference_example}.

% The numbering of figures, tables, equations and pages has been reset to start from S1, as in
% \begin{equation}
% 	\cos(2\theta) = \cos^2\theta - \sin^2\theta.
% 	\label{eq:sup_example} % Use a logical label
% \end{equation}

% \subsubsection*{Example supplement heading}

% The two main sections of the supplement can be split up using headings.

\subsubsection*{Robot as a mobile entity}
We use a Mecabot Pro omnidirectional mobile robot (Roboworks, Australia) as the physical platform for our MM-LLM experiment, using its integrated sensor array as the primary information source. This research-grade robot operates on the Robotic Operating System (ROS) framework and features a sensor suite including LiDAR, encoders, an RGB-D camera, and an Inertial Measurement Unit (IMU). The robot measures 541 × 225.5 × 581 mm (length × height × width), weighs 10.8 kg, and can achieve a maximum speed of 1.83 m/s.

Thanks to its versatile sensors, high maneuverability, and strong exploration capabilities, the Mecabot Pro is well-suited for this experiment. Its various sensors are responsible for analyzing the surrounding environment, with the collected data being published via ROS2 topics. This results in a predefined, structured data format that is easy to extract and analyze.

This project uses ROS2 Humble Edition, which is compatible with Ubuntu 22.04.

\subsubsection*{Perceptual Framework: robot's multimodal sensory array}
The sensors to be used for the experiments conducted in this study are described below:

\textit{Encoders}: mounted on each of the robot's four wheels. With a resolution of 500 lines per revolution, the encoders provide granular motion data critical for precise position, speed, and orientation control. Encoder measurements are continuously published to the ROS2 \textit{/odometry/filtered} topic, delivering real-time updates on the robot's position, linear velocity, and orientation parameters.

\textit{IMU}: integrated within the Mecabot's STM32 microcontroller (STMicroelectronics, Switzerland). This sensor continuously monitors the robot's linear acceleration across all axes. All IMU measurements are published in real-time to the ROS2 topic \textit{/mobile\_base/sensors/imu\_data}.

\textit{LiDAR}: model N10 (Leishen, China) installed on top of the Mecabot Pro. It offers 360$^\circ$ coverage of the robot's environment, a 30 m detection range, a 12 Hz scanning frequency, and an angular increment between measurements of 0.68$^\circ$. In one complete LiDAR cycle, 529 distances are measured. This generates an excessive amount of data, which can lead to processing issues and LLM saturation. To address this, the data is simplified by dividing the area surrounding the robot into eight regions, similar to a compass rose (front, right, left, rear, front-right, front-left, rear-right, rear-left). For each region, only the closest obstacle distance is stored, ensuring that the most relevant information is retained. The information is published in the ROS2 \textit{/scan} topic. 

\textit{RGB-D camera}: embedded at the top of the robot's structure. It has a resolution of 640 x 480 px. The images captured by the camera are published in the ROS2 topic \textit{/camera/color/image\_raw}. This camera can also measure distances to different points in the image by generating a point cloud. However, the distance information is discarded due to the large volume of data involved, which could cause issues when processed by the LLM. Additionally, this information is highly similar to what is provided by the LiDAR sensor.

\subsubsection*{Raw Sensory Data Processing and Representation}
Each sensor operates at its own native resolution and sampling frequency. To unify formats and structure the data in a form suitable for analysis by the MM-LLM, we define a unified JSON representation. Each JSON instance aggregates data from the four sensory sources employed in the system: odometry, RGB camera, LiDAR, and IMU.

To ensure temporal coordination across modalities, all sensor readings included in a single JSON instance are condensed within a 50~ms window using a nearest-neighbor temporal matching strategy. To reduce data volume and avoid unnecessary model overload, the data stream is downsampled to 1~Hz, collecting one synchronized sample per second, and all numerical values are truncated to one decimal place.

Each JSON instance contains five fields:
\begin{itemize}
    \item \textit{Timestamp}: expressed in seconds with one decimal. ROS~2 provides timestamps in absolute UNIX epoch time (January~1,~1970,~00:00:00~UTC); these values are re-referenced to start at 0.0~s for the first sample.
    \item \textit{Odometry}: subdivided into position $(x,y,z)$, orientation (quaternion), linear velocity $(x,y,z)$, and angular velocity $(x,y,z)$, all expressed in International System (SI) units.
    \item \textit{IMU}: linear acceleration $(x,y,z)$ expressed in SI units.
    \item \textit{Image}: RGB image with a resolution of $480 \times 640$ pixels.
    \item \textit{Scan}: distance to the closest obstacle in eight directions (front, front-right, right, rear-right, rear, rear-left, left, and front-left), expressed in SI units.
\end{itemize}

\subsubsection*{LLM output}

The MM-LLM is provided at each instance with the JSON object containing all the sensory data. To prevent biasing the model’s
self-identification process, we deliberately employ ambiguous terminology, replac-
ing technical robotics terms with more generic alternatives (e.g., “sensors” become
“sources of information”,“encoders” become “position, linear velocity and orienta-
tion”, “LiDAR” become “proximity to obstacles”; “IMU” by “linear acceleration”
and “RGB-D camera” become “image”). 

For analyzing the results and predictions generated by the MM-LLM, we utilize JSON format because of its simplicity in handling and interpretation. The language model is specifically instructed to structure its responses as a JSON document containing four distinct fields:
\begin{itemize}
    \item Dimensions: estimate of its dimensions (length x height x width) in meters. 
    \item Movement: an explanation of the type of movement it performs to move around its environment.
    \item Entity: what type of individual it is, for example, a robot, a car or a human being. 
    \item Environment: description of the information extracted from the image. 
\end{itemize}

\subsubsection*{Memory storage and past predictions}
When processing the robot's sensory data, enabling the MM-LLM to refine its estimates over time requires an effective strategy for information storage and retrieval. To address this challenge, our approach implements an iterative memory development process. At each iteration, the MM-LLM generates a comprehensive summary integrating its previous estimates with new sensor-derived perceptions. This summary is stored and subsequently utilized as contextual information for the following iteration, creating a continuous feedback loop that enables progressive refinement of the model's understanding.

The MM-LLM receives the latest version of the past predictions and perception summary at each iteration, integrating them with the current sensory data. It then generates a response that considers both sources of information. This newly generated prediction subsequently serves as the updated summary for the next iteration, ensuring a continuous cycle of learning and refinement. 

\subsubsection*{Prompting}

We develop a structured prompt that guides the MM-LLM through four sequential phases. First, the model receives a generic introduction to its context and objectives—determining its identity, dimensions, movement modality and environment—using deliberately non-robotic language. Second, we list its four information sources (position/orientation/velocity, proximity to obstacles, linear acceleration and image) and provide access to an episodic summary of prior predictions. Third, we specify two core tasks—scene analysis and self-localization—to frame subsequent questions. Fourth, we enforce a JSON response format with fields for dimensions, movement, identity and image description, illustrated by an implausible example (e.g. a ``blue flying whale'') to test abstraction and formatting compliance.

At the end of the prompt, we append operational constraints, that require the model to rely solely on current sensory data and memory summaries, to maintain continuity, to operate autonomously and to always provide an estimate (never ``unknown'') even under limited information. If visual data are unavailable, the image field must read ``No visual information available''. These rules ensure systematic, iterative self-predictions and allow us to evaluate the MM-LLM’s sensorimotor reasoning under uncertainty. The complete prompt can be found in the code of the project. 

% \subsubsection*{Evaluation system: LLM-as-a-judge}
% We develop four distinct rubrics corresponding to each field in the system's JSON (dimensions, movement, entity and environment). Each rubric provides explicit guidance to the evaluating LLM using a rating scale from zero (worst response) to five (best response). For each of the grades from zero to five, a very detailed description of the conditions that a response must satisfy to acquire that score is assigned. Gemini 2.0 Flash serves as our evaluation LLM. The complete rubrics can be found in the code of the project. 

\subsubsection*{Evaluation system: LLM-as-a-Judge}
\begin{figure}[ht]
    \centering
    \includegraphics[width=0.8\linewidth]{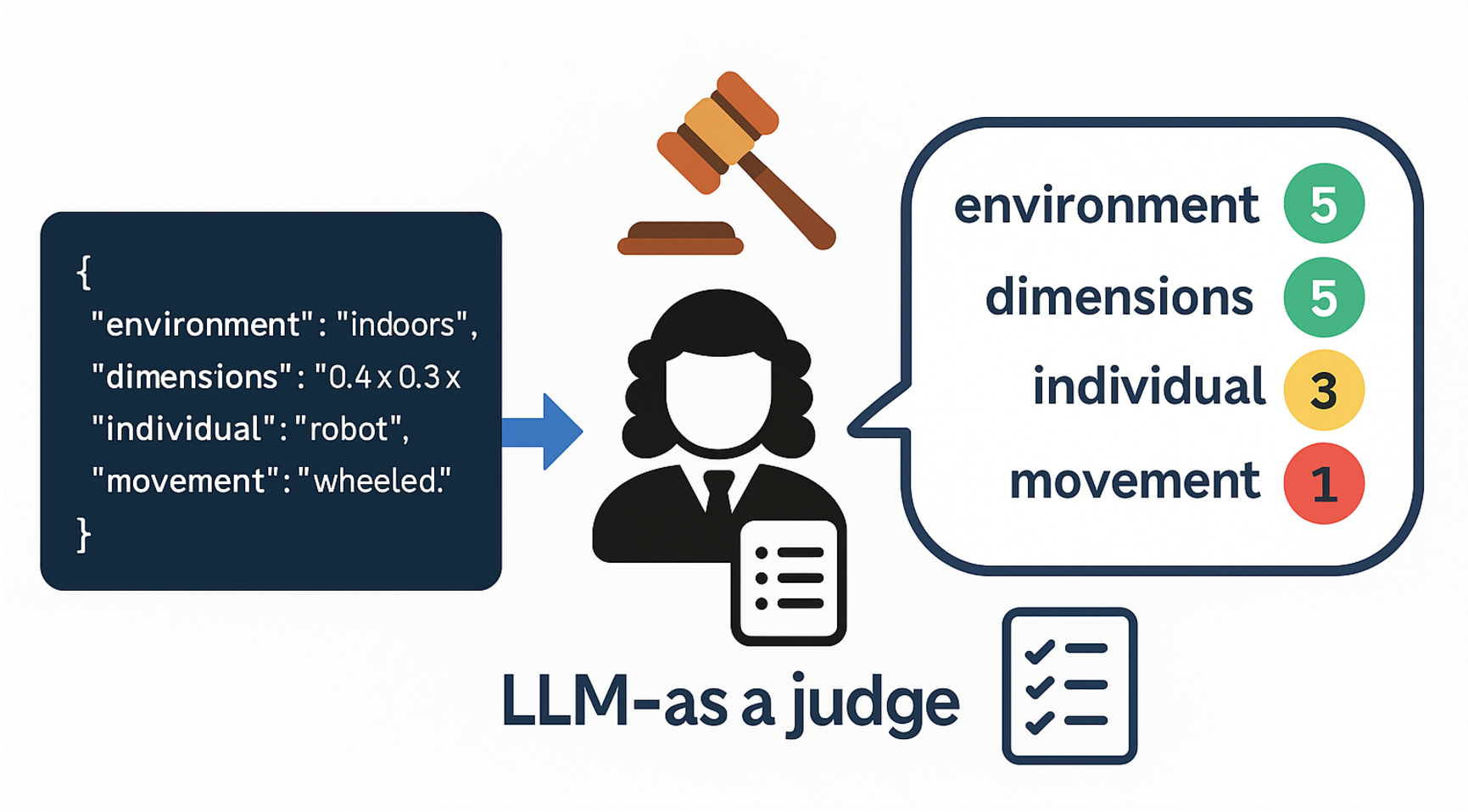}
    \caption{
Overview of the evaluation framework used to assess system outputs. Raw sensorimotor data are processed by the embodied MM-LLM to generate structured self-descriptions encoded in JSON format. These outputs are then evaluated by an independent LLM-as-a-Judge using dimension-specific rubrics and a shared ordinal scale from 0 to 5. The judge operates exclusively as an evaluation module, mapping qualitative predictions to rubric-defined scores for each self-perception dimension, enabling consistent and comparable quantitative analysis across experimental conditions.
}
    \label{fig:LLMAsAJudge}
\end{figure}
We evaluate the system outputs using an LLM-as-a-Judge framework, a methodology increasingly adopted for the structured assessment of qualitative and semantic model predictions. Four distinct rubrics are defined, corresponding to the four evaluated dimensions encoded in the system’s JSON output: \textit{Dimensions}, \textit{Movement}, \textit{Entity}, and \textit{Environment}. Each rubric provides explicit evaluation criteria and a shared ordinal rating scale ranging from 0 (completely erroneous or misconceived prediction) to 5 (outstanding and fully consistent prediction).

For each score level between 0 and 5, the rubrics specify detailed and mutually exclusive conditions that a response must satisfy in order to be assigned that score. This design enforces consistency across evaluations and reduces ambiguity in the interpretation of qualitative outputs. The same rubric structure is applied across all experimental conditions, enabling direct comparison between full-sensing and ablation scenarios.

Gemini~2.0~Flash is employed as the evaluation LLM. Its role is strictly evaluative: the model does not participate in generation, decision-making, or control, but solely maps system-generated textual descriptions to rubric-defined ordinal scores. This approach mirrors expert-based evaluation protocols commonly used in psychometrics and human–computer interaction, where latent constructs are operationalized through structured judgment rather than direct measurement.

The evaluation process is illustrated in Fig.~\ref{fig:LLMAsAJudge}, which summarizes the information flow from raw system outputs to rubric-based scoring. The complete rubric definitions, including the full description of each score level for all dimensions, are provided in the project code to ensure transparency and reproducibility.

\subsubsection*{Structural Equation Modeling}

We apply Structural Equation Modeling (SEM) to quantify the directional influences of sensory integration and memory on emergent self-recognition constructs. Our sample comprises $n=657$ iterations, each yielding four rubric-based scores that serve as observed endogenous indicators. These indicators are collected in the vector $\mathbf{Y}$ (Eq.~\ref{eq:Y}), corresponding to the rubric scores for \textit{Dimensions}, \textit{Movement}, \textit{Image}, and \textit{Individual} self-recognition.

Exogenous variables $\boldsymbol{\xi}$ (Eq.~\ref{eq:xi}) include Z-score–normalized proprioceptive and inertial measurements—position, orientation, linear velocity, and linear acceleration—together with two binary indicators encoding the presence of episodic memory and visual input. These variables represent the externally provided sensory and memory-related signals that drive the emergence of higher-level self-related constructs. LiDAR inputs are excluded from the SEM due to high redundancy and negligible contribution observed in preliminary analyses.

Latent endogenous constructs $\boldsymbol{\eta}$ (Eq.~\ref{eq:eta_vector}) model internal self-related representations inferred from the observed indicators. These constructs include \textit{Past–Present Memory}, \textit{Dimension Awareness}, \textit{Movement Awareness}, \textit{Environmental Awareness}, and \textit{Self-Identification}, which together define a hierarchical organization of self-recognition processes.

The SEM is specified in two stages. First, the measurement model relates the observed indicators to the latent constructs:
\[
  \mathbf{Y} = \Lambda_{y}\,\boldsymbol{\eta} + \varepsilon.
\]
Second, the structural model captures the directed dependencies among latent constructs and their modulation by exogenous variables:
\[
  \boldsymbol{\eta} = B\,\boldsymbol{\eta} + \Gamma\,\boldsymbol{\xi} + \zeta.
\]

Structural relations are assessed using standardized path coefficients $\beta^*$ (significant at $p<0.05$), and overall model fit is evaluated using the comparative fit index (CFI), Tucker–Lewis index (TLI), and root mean square error of approximation (RMSE). The final model achieves CFI~=~0.97, TLI~=~0.95, and RMSE~=~0.08, indicating a well-fitting hierarchical representation of sensorimotor self-recognition.

\begin{equation}
\mathbf{Y} =
    \begin{bmatrix}
        \text{rubric\_Dimensions} & 
        \text{rubric\_Movement} & 
        \text{rubric\_Image} & 
        \text{rubric\_Individual}
    \end{bmatrix}^T
\label{eq:Y}
\end{equation}

\begin{equation}
    \boldsymbol{\xi} =
    \begin{bmatrix}
        \text{Memory} &
        \text{Image} &
        \text{Position} &
        \text{Orientation} &
        \text{Velocity} &
        \text{Acceleration}
    \end{bmatrix}^T
    \label{eq:xi}
\end{equation}

\begin{equation}
    \boldsymbol{\eta} =
    \begin{bmatrix}
        \text{PastPresentMemory} \\
        \text{DimensionAwareness} \\
        \text{MovementAwareness} \\
        \text{EnvironmentalAwareness} \\
        \text{SelfIdentification}
    \end{bmatrix}
    \label{eq:eta_vector}
\end{equation}

\end{document}